\documentclass[a4paper,conference]{IEEEtran}
\usepackage{amsmath,amssymb,amsfonts}
\usepackage{algorithmic}
\usepackage{mathtools}
\usepackage{graphicx}
\usepackage{times}
\usepackage{bm}
\usepackage{cite}
\usepackage{graphicx}
\usepackage[hidelinks]{hyperref}
\begin{document}
%
\title{Explanation-Guided Training for Cross-Domain Few-Shot Classification}




%
\author{\IEEEauthorblockN{Jiamei Sun\IEEEauthorrefmark{1},
Sebastian Lapuschkin\IEEEauthorrefmark{2}\,
Wojciech Samek\IEEEauthorrefmark{2},
Yunqing Zhao\IEEEauthorrefmark{1},\\
Ngai-Man Cheung\IEEEauthorrefmark{1}, and
Alexander Binder\IEEEauthorrefmark{1}}
\IEEEauthorblockA{\IEEEauthorrefmark{1}Information System of Technology and Design, Singapore University of Technology and Design
}
\IEEEauthorblockA{\IEEEauthorrefmark{2}Department of Video Coding \& Analytics, Fraunhofer Heinrich Hertz Institute, Berlin, Germany}
\IEEEauthorblockA{Email: jiamei\_sun@mymail.sutd.edu.sg;sebastian.lapuschkin@hhi.fraunhofer.de;wojciech.samek@hhi.fraunhofer.de\\yunqing\_zhao@mymail.sutd.edu.sg;ngaiman\_cheung@sutd.edu.sg;alexander\_binder@sutd.edu.sg}
}


\maketitle

\begin{abstract}
Cross-domain few-shot classification task (CD-FSC) combines few-shot classification with the requirement to generalize across domains represented by datasets. This setup faces challenges originating from the limited labeled data in each class and, additionally, from the domain shift between training and test sets. 
In this paper, we introduce a novel training approach for existing FSC models. It leverages on the explanation scores, obtained from existing explanation methods when applied to the predictions of FSC models, computed for intermediate feature maps of the models. Firstly, we tailor the layer-wise relevance propagation (LRP) method to explain the predictions of FSC models. Secondly, we develop a model-agnostic explanation-guided training strategy that dynamically finds and emphasizes the features which are important for the predictions. Our contribution does not target a novel explanation method but lies in a novel application of explanations for the training phase. We show that explanation-guided training effectively improves the model generalization. We observe improved accuracy for three different FSC models: RelationNet, cross attention network, and a graph neural network-based formulation, on five few-shot learning datasets: miniImagenet, CUB, Cars, Places, and Plantae. The source code is available  \url{https://github.com/SunJiamei/few-shot-lrp-guided}
\end{abstract}

\IEEEpeerreviewmaketitle

\section{Introduction}
\label{sec:introduction}
Human beings can recognize new objects after seeing only a few examples. However, common image classification models require large amounts of labeled samples for training or fine-tuning.
To address this issue, few-shot classification (FSC) aims at generalization to new categories with a few training samples 
\cite{MatchNet:vinyals2016matching, MAML:finn2017model, PROTO:snell2017prototypical, RN:sung2018learning, FEWGNN:garcia2018fewshot,LEO:rusu2018metalearning, MTL:sun2019meta, CAN:hou2019cross}. This is relevant for setups, in which humans annotate a few examples of novel categories after model deployment, not present in the originally trained model.
FSC models are commonly evaluated using test data from the same domain as the training dataset. Lately, \cite{ACLOSERLOOK:chen2018a} stated that the current FSC methods meet difficulties in cases exhibiting domain shifts between the training data (source domain) and the test data (target domain). 
For example, people can recognize different kinds of birds and plants with a few examples of each category. Contrasting, existing FSC models trained on the bird domain may not accurately recognize various kinds of plants, which is demonstrated in \cite{ACLOSERLOOK:chen2018a, FeaturewiseTranslayer:tseng2020cross}.
The cross-domain few-shot classification is a more challenging and more useful task.

Tackling the domain shift problem requires additional efforts to avoid overfitting to the source domain. 
A recent work addresses the domain shift issue by learning a noise distribution for intermediate layers in the feature encoder \cite{FeaturewiseTranslayer:tseng2020cross}. Other approaches rely on adding batch spectral regularization over the encoded image features \cite{BSR:liu2020feature} and employing novel losses \cite{LMM:yeh2020large, ACLOSERLOOK:chen2018a}. This paper proposes a novel approach for improving CD-FSC models from a different perspective: we leverage on explanations computed for intermediate feature maps of FSC models to guide the model to learn better feature representations. For explanations, we refer to methods such as gradient- or Shapley-type methods, LRP\cite{LRP:bach2015pixel} or LIME\cite{LIME:ribeiro2016should} that compute a score for every dimension of a feature map, denoting the importance to the final prediction.

Although a large number of explanation methods have contributed substantial progress to the field of explaining model predictions \cite{GRADIENT:simonyan2013deep,IntegratedGradient:sundararajan2017axiomatic,LRP:bach2015pixel,GRADCAM:selvaraju2017grad, LIME:ribeiro2016should, DEEPTAYLOR:Lmontavon2017explaining, DEEPLIFT:shrikumar2017learning, PATTERN:kindermans2017learning}, they are usually applied in the testing phase, and frequently, do not consider the use cases of explanations. Some known use cases are the audit of predictions \cite{ASSESS:Lapuschkin2019}, explanation-weighed document representations that are more comprehensive\cite{LRPreweight:arras2017relevant}, and identification of biases in datasets \cite{GRADCAM:selvaraju2017grad}. We will add a new use case for explanations during the training phase, and consider whether the explanations are suitable to improve model performance in cross-domain few-shot classification. 

Many explanation methods \cite{LRP:bach2015pixel,GRADIENT:simonyan2013deep, GRADCAM:selvaraju2017grad} explain predictions on a per-sample basis. With a target label and an input sample, these explanation methods assign scores to each neuron of every feature map within the model. These scores are related to the importance of a neuron to the target label. Explanations are generated usually with a modified backward-pass and require no additional trainable parameters inside the model. In this paper, we study whether the explanation scores of intermediate feature maps can be employed to improve model generality in the few-shot classification, which is still a novel question. 

\begin{figure}[tb]
    \centering
    \includegraphics[width = 0.45\textwidth]{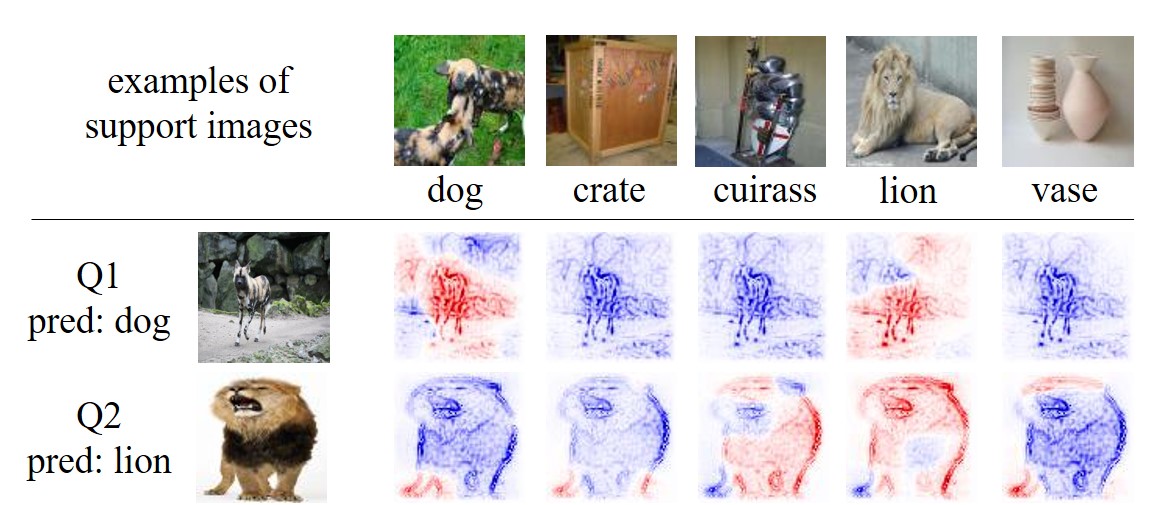}
    \caption{LRP explanation heatmaps of the input image with 5 target labels. The experiment model is a RelationNet trained on miniImagenet under the 5-way 5-shot setting. The first row illustrates some examples of the support images. The other two rows show the explanation heatmaps of two query images, Q1: \textit{African hunting dog} (denoted as \textit{dog}) and Q2: \textit{lion}. Both images are correctly predicted and the heatmaps are generated using different target labels. Red pixels indicate positive LRP relevance score and blue indicates negative. The strength of the color corresponds to the value of the LRP relevance scores.}
    \label{fig:introLRPexample}
\end{figure}
Concretely, we adapt LRP-type explanations\cite{LRP:bach2015pixel} to FSC models. LRP has been used to explain convolutional neural networks (CNN)\cite{LRP:bach2015pixel}, recurrent neural networks (RNN)\cite{LRP-LSTM:arras2017explaining}, graph neural networks (GNN)\cite{GNNLRP:schnake2020xai}, and clustering models\cite{Explainkmeans:kauffmann2019clustering}. 
It backpropagates the relevance score of a target label through the neural network and assigns the relevance scores to the neurons within the network. The sign and the amplitude of LRP relevance scores reflect the contribution of a neuron to the prediction, as shown in Figure \ref{fig:introLRPexample}. Relying on this property, we propose ``explanation-guided training'' for FSC models. The LRP relevance scores of intermediate feature maps are employed as weights and used to construct LRP-weighted feature maps. This step emphasizes the feature dimensions, which are more relevant to the model prediction, and downscales the less relevant ones. 
The LRP-weighted features are then fed into the network to guide the training. 
Since LRP explanations are calculated for each sample-label pair separately, our explanation-guided training adds a label-dependent feature weighting mechanism during training. We will show that this mechanism can reduce overfitting to the source domain. 
We remark that the principles used for explanation-guided training strategy are model-agnostic and can be combined with other CD-FSC methods such as the learned feature-wise transformation (LFT) \cite{FeaturewiseTranslayer:tseng2020cross} and other explanation methods. 
The main contributions of this paper are described as follows.
\begin{itemize}
\item We derive explanations for FSC models using LRP.
\item We investigate the potential of improving model performance using explanations in the training phase under few-shot settings.
\item We propose an explanation-guided training strategy to tackle the domain shift problem in FSC. 
\item We conduct experiments to show that the explanation-guided training strategy improves the model generalization for a number of FSC models and datasets. 
\item We combine our explanation-guided training strategy with another recent approach, LFT \cite{FeaturewiseTranslayer:tseng2020cross}, which shares with our approach the property of being applicable on top of existing models, and observe a synergy of these two methods further improves the performance.
\end{itemize}


\section{Related Work}
\label{sec:related work}
\subsection{Few-shot Classification Methods}
Optimization-based and metric-based approaches constitute two prominent directions in few-shot learning. The former one is learning initialization parameters that can be quickly adapted to new categories \cite{MAML:finn2017model,Reptile:nichol2018reptile,LEO:rusu2018metalearning,MTL:sun2019meta} or designing a meta-optimizer that learns how to update the model parameters\cite{LSTMoptimizer:ravi2016optimization, MANN:santoro2016meta, SNAIL:mishra2018a, METANETWORK:munkhdalai2017meta}. Metric-based methods learn a distance metric to compare the support and query samples and classify the query image to the closest category\cite{MatchNet:vinyals2016matching, RN:sung2018learning, PROTO:snell2017prototypical, FEWGNN:garcia2018fewshot, TPN:liu2018learning, SUBSPACE:devos2019subspace, CAN:hou2019cross}. Other approaches are noteworthy. \cite{TADAM:oreshkin2018tadam, SIMPLECANPS:bateni2020improved} design and add task-conditional layers to the model. \cite{DYNAMIC:gidaris2018dynamic, ImprintedWeight:qi2018low, GCW:gidaris2019generating} dynamically update the classifier weight for new categories. \cite{MARGINLOSS:li2020boosting, CROSSMODAL:xing2019adaptive} combine multiple modal information such as the word embedding of the class label. \cite{Imaginary:wang2018low} augments the training data by hallucinating new samples. \cite{LST:li2019learning, SEMISUP:ren2018meta} leverage unlabeled training samples and semi-supervising training strategy. \cite{SELFSupervise:Gidaris_2019_ICCV} equips the model with a self-supervision mechanism. However, recent research discussed that existing FSC methods may meet difficulties with domain shift, a more challenging and practical problem\cite{ACLOSERLOOK:chen2018a}. 

\subsection{Cross-domain Few-shot Classification Methods}
It is common to develop cross-domain few-shot classification methods from existing FSC methods.
LFT\cite{FeaturewiseTranslayer:tseng2020cross} learns a noise distribution and adds the noise to intermediate feature maps to generate more diverse features during training and improve the model generality. In the recent CVPR Cross-Domain Few-Shot Learning challenges \cite{CDFSLCVPR:guo2019new, BSR:liu2020feature} ensembled multiple feature encoders and employed batch spectral regularization over the image features for each encoder. Batch spectral regularization penalizes the singular values of the feature matrix within a batch so that the learned features maintain similar spectra across domains. \cite{MetaFT:cai2020cross} combined the first-order MAML\cite{MAML:finn2017model} and the GNN metric-based method \cite{FEWGNN:garcia2018fewshot}.  
\cite{LMM:yeh2020large} applied a prototypical triplet loss to increase the inter-class distance and a large margin cosine loss to minimize the intra-class distance, which is also studied by \cite{ACLOSERLOOK:chen2018a} that reducing intra-class variation benefits FSC, especially for shallow image feature encoders. 
In our approach, we do not introduce more parameters like \cite{FeaturewiseTranslayer:tseng2020cross}. We are similar to \cite{BSR:liu2020feature} and \cite{LMM:yeh2020large} in adding constraints on the image features. We are different in using LRP-weighted features to guide the model to dynamically correct itself for each instance instead of penalizing feature statistics over a batch. The LRP-weighting idea has been used to generate more comprehensive document representations \cite{LRPreweight:arras2017relevant}. We are different from \cite{LRPreweight:arras2017relevant} that the re-weighting strategy is embedding into the training phase to improve the model.

\subsection{Explanation for Few-shot Classification Models}
There exist explanation methods for deep neural networks (DNN) \cite{LRP:bach2015pixel, DEEPTAYLOR:Lmontavon2017explaining, GRADIENT:simonyan2013deep, GuidebackPropagation:springenberg2014striving, GRADCAM:selvaraju2017grad, GNNLRP:schnake2020xai} that can be adapted to FSC models, since many FSC models adopt CNN to encode image features and many metric-based methods also adopt DNN to learn the distance metric \cite{RN:sung2018learning, FEWGNN:garcia2018fewshot, TPN:liu2018learning}. For FSC models that use non-parametric distance metrics, we refer to \cite{Explainkmeans:kauffmann2019clustering} that transforms various K-means classifiers into neural net structures and then applies LRP to obtain explanations. 
In this paper, we have chosen LRP due to its reasonable performance \cite{COMPARISON:poerner2018evaluating}, our understanding of its hyperparameters, and its reasonable speed compared to LIME or some theoretically equally well-motivated but exhaustive Shapley-type approaches. While using other explanation methods among the faster ones would be possible, this would not change the qualitative message of this paper regarding the applicability of explanation methods for few-shot training. The results here are meant as a case for explanation methods in general, even when they are demonstrated for one approach. 

\section{Explanation-Guided Training}
\label{sec:EGT}
\begin{figure}[tb]
    \centering
    \includegraphics[width = 0.45\textwidth]{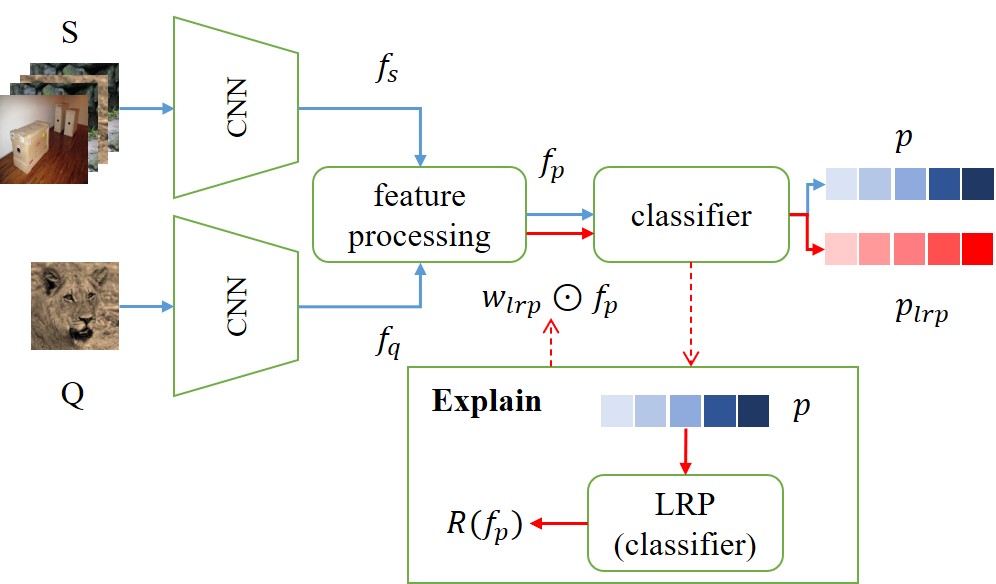}
    \caption{Explanation-guided training. Blue paths denote the conventional FSC training. The red paths are originating from the explanation  method. They are added after one step following the blue paths. The support samples $\mathcal{S}$ and the query sample $\mathcal{Q}$ are fed into an image encoder to obtain features $f_s$ and $f_q$, which are further processed by a \textit{feature processing} module. The output of \textit{feature processing} $f_p$ is fed into a classifier to make predictions. Both the \textit{feature processing} and \textit{classifier} modules vary across different FSC methods. The \textbf{Explain} block explains the model prediction $p$ and generate the explanations for $f_p$, denoted as $R(f_p)$, which are used to calculate the LRP weight $w_{lrp}$. The LRP-weighted feature $w_{lrp} \odot f_p$ is fed into the classifier resulting in the updated prediction $p_{lrp}$.}
    \label{fig:EGT_frame}
\end{figure}

Before presenting our explanation-guided training, we first introduce the cross-domain few-shot learning task and the notations.
For a K-way N-shot task, denoted as an episode, we are given a support set $\mathcal{S}=\{(x_s, y_s)\}^{K*N}_{s=1}$ containing $K$ classes and $N$ labeled samples per class for training and a query set $\mathcal{Q}=\{(x_q, y_q)\}_{q=1}^{n_q}$ from the same classes as $\mathcal{S}$ for testing. The CD-FSC task is to train an FSC model using episodes $\{\mathcal{S}_i, \mathcal{Q}_i\}$ randomly sampled from a base domain $\mathcal{D}_{seen}$ and test the model with episodes sampled from an unseen domain $\mathcal{D}_{unseen}$.
We consider FSC models that can be outlined as Figure \ref{fig:EGT_frame} in our study. This includes many metric-based FSC models. 

The support set $\mathcal{S}$ and query set $\mathcal{Q}$ are encoded by a CNN \cite{RN:sung2018learning, CAN:hou2019cross}, possibly with augmented layers \cite{TADAM:oreshkin2018tadam, FeaturewiseTranslayer:tseng2020cross} to obtain the support image features $f_s$ and the query image features $f_q$. $f_s$ and $f_q$ are further processed before classification, for example,  \cite{RN:sung2018learning} simply averages the $f_s$ over classes and concatenate the averaged class representations pairwise with $f_q$, \cite{CAN:hou2019cross} designs an attention module and generate the attention-weighted support and query image features, \cite{TPN:liu2018learning} applies GNN on $f_s$ and $f_q$ to obtain graph structured features. 
The processed features are fed into a classifier for predictions. The classifier can be cosine similarity\cite{CAN:hou2019cross}, Euclidean distances\cite{PROTO:snell2017prototypical}, Mahalanobis distance\cite{SIMPLECANPS:bateni2020improved}, or neural nets\cite{RN:sung2018learning, FEWGNN:garcia2018fewshot}.

Explanation-guided training for FSC models involves the following steps. For each training episode:

\textbf{Step1}: One forward-pass through the model and obtain the prediction $p$, illustrated as the blue path in Figure \ref{fig:EGT_frame}.

\textbf{Step2}: \textbf{Explaining the classifier}. We initialize the LRP relevance for each label and apply LRP to explain the classifier. We can obtain the relevance of the classifier input $R(f_p)$, illustrated as the \textbf{Explain} block. 

For FSC models that implement a neural network as the classifier, the relevance scores for each label can be initialized with their logits. For the models using non-parametric distance measures such as cosine similarity and Euclidean distance, the predicted scores are positive for all labels, which will result in similar explanations.
For such cases, we refer to the logit function in \cite{Explainkmeans:kauffmann2019clustering} to initialize the relevance scores. Taking the cosine similarity as an example, we first calculate the probability for each class using the exponential function via equation \eqref{equ:softmax} \footnote{For distance measures such as Euclidean distance, we need to use the negative distance to replace the similarity metric.}.
\begin{equation}
        P(y_c|f_p) = \frac{exp(\beta \cdot cs_c(f_p))}{\sum_{k=1}^K exp(\beta \cdot cs_k(f_p))} \label{equ:softmax}
\end{equation}
$cs_k(\cdot)$ means the cosine similarity between a query sample and class $k$. $f_p$ is the processed feature fed to the classifier. $\beta$ is a constant scale parameter to strengthen the highest probability. Using the  probability defined above, the relevance score of class $c$ is defined as:
\begin{equation}
    R_c = log\left(\frac{P(y_c|f_p)}{1-P(y_c|f_p)} (K-1)\right) \label{equ:logits}
\end{equation}
$R_c,c=1\dots K$ is positive when the $P(y_c|f_p)$ is larger than $1/K$. In other words, the class label whose probability is larger than the random guessing probability receives a positive relevance score. 
With the relevance score of each target label $R_c$, standard LRP is applicable to backpropagate $R_c$ through the classifier to generate the explanations. 

Consider the forward pass from layer $l$ to layer $l+1$ as:
\begin{equation}
    \begin{aligned}
        y^{l+1}_j &=  \sum_i w_{ij}z^l_i + b_j\\
        z^{l+1}_j &=f(y^{l+1}_j)
    \end{aligned}
\end{equation}
where $i$ and $j$ are the indices of neurons in $l^{th}$ and $l+1^{th}$ layer, $f(\cdot)$ is an activation function. Let $R(\cdot)$ denote the relevance of a neuron and $R_{i\leftarrow j}$ denote the relevance attribution from $z^{l+1}_j$ to $z^l_i$.  We rely on two established LRP backpropagation mechanisms here, the $LRP_{\epsilon}$-rule and the $LRP_{\alpha}$-rule \cite{LRP:bach2015pixel}.
\begin{enumerate}
    \item $LRP_{\epsilon}$-rule
\begin{equation}
R_{i \leftarrow j} = R(z^{l+1}_j)\frac{z^l_iw_{ij}}{y^{l+1}_j + \epsilon \odot \mathrm{sign}(y^{l+1}_j)}
\end{equation}
$\epsilon$ is a small positive number and $\epsilon \odot  \mathrm{sign}(y^{l+1}_j)$ guarantees safe division.
\item $LRP_{\alpha}$-rule
\begin{equation}
    R_{i \leftarrow j} = R(z^{l+1}_j)\left(\alpha\frac{(z^l_iw_{ij})^+}{(y^{l+1}_j)^+}-(\alpha-1)\frac{(z^l_iw_{ij})^-}{(y^{l+1}_j)^-}\right)
\end{equation}
where $\alpha \geqslant 1$ controls the ratio of positive relevance to backpropagate. 
$(\ast)^+=\max(\ast,0)$, $(\ast)^-=\min(\ast,0)$.  
\end{enumerate}
The relevance of $z^l_i$ is the summation of all the relevance attribution flowing to it.
\begin{equation}
    R(z^l_i)=\sum_jR_{i\leftarrow j}
\end{equation}
We adopt the $LRP_{\epsilon}$-rule for linear layers and the $LRP_{\alpha}$-rule for convolutional layers to obtain $R(f_p)$, which is the suggested setting for explaining CNNs in \cite{SebasIJCNN2020:kohlbrenner2019towards}.
$R(f_p)$ is normalized by its maximal absolute value.

\textbf{Step3}: \textbf{LRP-weighted features}.
To emphasize the features which are more relevant to the prediction and downscale the less relevant ones, we define the LRP weights and the LRP-weighted features as
\begin{align}
    w_{lrp} &= 1 + R(f_p)\\
    f_{p-lrp} &= w_{lrp} \odot f_p
\end{align}
where $\odot$ is the element-wise product. Note that $R(f_p) \in [-1,1]$ after normalization, thus $w_{lrp}$ magnifies the features with positive relevance scores and downscales those with negative relevance scores. The maximal feature scaling after weighting with $w_{lrp}$ is $2$.

\textbf{Step4}: Finally, we forward the LRP-weighted features to the classifier to generate the explanation-guided predictions $p_{lrp}$. The objective function merges both the model prediction $p$ and the explanation-guided prediction $p_{lrp}$.
    \begin{equation}
        \mathcal{L} = \xi \mathcal{L}_{ce}(y, p)+\lambda \mathcal{L}_{ce}(y, p_{lrp}) \label{equ:loss}
    \end{equation}
where $\mathcal{L}_{ce}$ is the cross entropy loss. $\xi$ and $\lambda$ are positive scalars that control how much information from $p$ and $p_{lrp}$ are used. In our experiment, $\xi$ and $\lambda$ are empirically adjusted for different FSC models. 
\section{Experiments}
\label{sec:experiment}
We evaluate the proposed explanation-guided training on RelationNet(RN)\cite{RN:sung2018learning} and two of the state-of-the-art models, cross attention network(CAN) \cite{CAN:hou2019cross} and GNN network \cite{FEWGNN:garcia2018fewshot}. The correspondence of the three FSC models to the framework in Figure \ref{fig:EGT_frame} is summarized in Table \ref{tab:RN_CAN_GNNmodules}. We will demonstrate that explanation-guided training improves the performance of the three models on 4 cross-domain test sets. 

Moreover, we also combine explanation-guided training with another approach, LFT \cite{FeaturewiseTranslayer:tseng2020cross}. We show that explanation-guided training is compatible with LFT and the combination further improves the performance. 

\subsection{Dataset and Model Preparation}
\begin{table}[tb]
    \centering
    \caption{The correspondence of RelationNet(RN), cross attention network(CAN), and graph neural network(GNN) to the framework in Figure \ref{fig:EGT_frame}.}
    \begin{tabular}{c c c c c}
    \hline
            & &  feature processing    & & classifier\\ \hline
        RN  & &  pairwise concatenation& & relation module\\
        CAN & &  cross attention module      & & cosine similarity \\ 
        GNN & &  \textit{fc} layer and concatenation& & graph neural network\\
        \hline
    \end{tabular}
    \label{tab:RN_CAN_GNNmodules}
\end{table}

Five datasets are used in our experiment including miniImagenet\cite{miniImageNet:vinyals2016matching}, CUB\cite{CUB:wah2011caltech}, Cars\cite{Cars:krause20133d}, Places\cite{Places:zhou2017places}, and Plantae\cite{Plantae:van2018inaturalist}, which are introduced in \cite{FeaturewiseTranslayer:tseng2020cross}. Each dataset consists of train/val/test splits. We choose miniImagenet as the $\mathcal{D}_{seen}$ and train the FSC models on the training set, validate the models on the validation set of miniImagenet, and adopt the test sets of the other four datasets for testing.

We use Resnet10\cite{Resnet:he2016deep} as the image encoder for RN and GNN, and Resnet12 for CAN model. The three models are trained under 5-way 5-shot and 5-way 1-shot settings.
The LRP parameters are $\alpha=1, \epsilon=0.001$ for all the experiments, following the suggestions in \cite{SebasIJCNN2020:kohlbrenner2019towards}.

We experimented with varying $\xi$ and $\lambda$ in eq\eqref{equ:loss} and observed that, for the classifiers with trainable parameters such as RN and GNN, fully relying on $\mathcal{L}_{ce}(y, p_{lrp})$ ($\xi = 0$) makes the model hard to converge and only gain marginal improvement, while it works well for the non-parametric classifier such as the cosine similarity in CAN. The reason is that the explanations for a poor classifier make less sense and will distract the parameters of the classifier from the beginning, especially when there are fewer shots (smaller N). Thus, we combine the $\mathcal{L}_{ce}(y, p)$ to stable the training and increase the contribution of $\mathcal{L}_{ce}(y, p)$ for 1-shot setting. In the experiments using RN and GNN, we set $\xi=1, \lambda=0.5$ for 5-way 1-shot setting and $\xi=1, \lambda=1$ for 5-way 5-shot setting. For the CAN model, we set $\beta$ in eq\eqref{equ:softmax} as 7, the same as the original model, and $\xi=0, \lambda=1$.

We follow the same implementation details as  \cite{FeaturewiseTranslayer:tseng2020cross}\footnote{\url{https://github.com/hytseng0509/CrossDomainFewShot}} and \cite{CAN:hou2019cross}\footnote{\url{https://github.com/blue-blue272/fewshot-CAN}} to train the RN, GNN, and CAN model. 
At test time, we evaluate the performance over 2000 randomly sampled episodes, with 16 query images per episode.
\begin{table*}[htb]
    \centering
        \caption{Evaluation of explanation-guided training on cross-domain datasets using RN and CAN. We report the average accuracy of over 2000 episodes with 95\% confidence intervals. The models are trained on the miniImagenet training set and tested on the test set of various domains. \textbf{LRP-} means explanation-guided training using LRP. \textbf{T }indicates transductive inference.}
        \begin{tabular}{c c c c c}
    \hline
      miniImagenet  & 1-shot                    &1-shot-T                   & 5-shot                     & 5-shot-T \\ \hline
      RN            & 58.31$\pm$0.47\%          &61.52$\pm$0.58\%           & 72.72$\pm$0.37\%           &73.64$\pm$0.40\%\\ 
      LRP-RN        & \textbf{60.06}$\pm$\textbf{0.47}\% &\textbf{62.65}$\pm$\textbf{0.56}\%  &\textbf{73.63}$\pm$\textbf{0.37}\%   &\textbf{74.67}$\pm$\textbf{0.39}\%\\ 
      CAN           & \textbf{64.66}$\pm$\textbf{0.48}\% &67.74$\pm$0.54\%           & 79.61$\pm$0.33\%           &80.34$\pm$0.35\%\\
      LRP-CAN       & 64.65$\pm$0.46\%          &\textbf{69.10}$\pm$\textbf{0.53}\%  & \textbf{80.89}$\pm$\textbf{0.32}\%  &\textbf{82.56}$\pm$\textbf{0.33}\%\\ 
      \hline\hline
      mini-CUB      & 1-shot                    &1-shot-T                   & 5-shot                     & 5-shot-T \\ \hline
      RN            & 41.98$\pm$0.41\%          &42.52$\pm$0.48\%           & 58.75$\pm$0.36\%           &59.10$\pm$0.42\%\\ 
      LRP-RN        & \textbf{42.44}$\pm$\textbf{0.41}\% &\textbf{42.88}$\pm$\textbf{0.48}\%  & \textbf{59.30}$\pm$\textbf{0.40}\%  &\textbf{59.22}$\pm$\textbf{0.42\%}\\ 
      CAN           & 44.91$\pm$0.41\%          &46.63$\pm$0.50\%           & 63.09$\pm$0.39\%           &62.09$\pm$0.43\%\\
      LRP-CAN       & \textbf{46.23}$\pm$\textbf{0.42}\% &\textbf{48.35}$\pm$\textbf{0.52}\%  & \textbf{66.58}$\pm$\textbf{0.39}\%  &\textbf{66.57}$\pm$\textbf{0.43}\%\\ 
     \hline\hline
     
      mini-Cars     & 1-shot                  &1-shot-T                 & 5-shot                   & 5-shot-T \\ \hline
      RN            & 29.32$\pm$0.34\%          &28.56$\pm$0.37\%           & 38.91$\pm$0.38\%           &37.45$\pm$0.40\%\\ 
      LRP-RN        & \textbf{29.65}$\pm$\textbf{0.33}\% &\textbf{29.61}$\pm$\textbf{0.37}\%  & \textbf{39.19}$\pm$\textbf{0.38}\%  &\textbf{38.31}$\pm$\textbf{0.39}\%\\ 
      CAN           & 31.44$\pm$0.35\%          &30.06$\pm$0.42\%           & 41.46$\pm$0.37\%           &40.17$\pm$0.40\%\\
      LRP-CAN       & \textbf{32.66}$\pm$\textbf{0.46}\% &\textbf{32.35}$\pm$\textbf{0.42}\%  &\textbf{43.86}$\pm$\textbf{0.38}\%  &\textbf{42.57}$\pm$\textbf{0.42}\%\\ 
      \hline\hline
      
      mini-Places   & 1-shot                  &1-shot-T                 & 5-shot                   & 5-shot-T \\ \hline
      RN            & \textbf{50.87}$\pm$\textbf{0.48}\% &\textbf{53.63}$\pm$\textbf{0.58}\%  & 66.47$\pm$0.41\%           &67.43$\pm$0.43\%\\ 
      LRP-RN        & 50.59$\pm$0.46\%          &53.07$\pm$0.57\%           & \textbf{66.90}$\pm$\textbf{0.40}\%  &\textbf{68.25}$\pm$\textbf{0.43}\%\\ 
      CAN           & 56.90$\pm$0.49\%          &60.70$\pm$0.58\%           & 72.94$\pm$0.38\%           &74.44$\pm$0.41\%\\
      LRP-CAN       & \textbf{56.96}$\pm$\textbf{0.48}\%&\textbf{61.60}$\pm$\textbf{0.58}\%   & \textbf{74.91}$\pm$\textbf{0.37}\%  &\textbf{76.90}$\pm$\textbf{0.39}\%\\ 
      \hline\hline
      
      mini-Plantae  & 1-shot                  &1-shot-T                 & 5-shot                   & 5-shot-T \\ \hline
      RN            & 33.53$\pm$0.36\%          &33.69$\pm$0.42\%           & 47.40$\pm$0.36\%           &46.51$\pm$0.40\%\\ 
      LRP-RN        & \textbf{34.80}$\pm$\textbf{0.37}\% &\textbf{34.54}$\pm$\textbf{0.42}\% & \textbf{48.09}$\pm$\textbf{0.35}\%  &\textbf{47.67}$\pm$\textbf{0.39}\%\\       
      CAN           & 36.57$\pm$0.37\%          &36.69$\pm$0.42\%           & 50.45$\pm$0.36\%           &48.67$\pm$0.40\%\\
      LRP-CAN       & \textbf{38.23}$\pm$\textbf{0.45}\% &\textbf{38.48}$\pm$\textbf{0.43}\% & \textbf{53.25}$\pm$\textbf{0.36}\% &\textbf{51.63}$\pm$\textbf{0.41}\%\\ 
      \hline
    \end{tabular}

    \label{tab:accuracy_single domain}
\end{table*}

\begin{table*}[tb]
    \centering
    \caption{Evaluation of explanation-guided training on cross-domain datasets using GNN. We report the average accuracy of over 2000 episodes with 95\% confidence intervals. The models are trained on the miniImagenet training set and tested on the test set of various domains. \textbf{LRP-} means explanation-guided training using LRP.}
    \begin{tabular}{c c c c c c}
    \hline
        5-way 1-shot  &miniImagenet             &  Cars           & Places         & CUB          & Plantae\\ \hline
        GNN           &64.47$\pm$0.55\%           & 30.97$\pm$0.37\%  &54.64$\pm$0.56\%  &46.76$\pm$0.50\%& 37.39$\pm$0.43\%\\
        LRP-GNN       &\textbf{65.03}$\pm$\textbf{0.54}\%  & \textbf{32.78}$\pm$\textbf{0.39}\%  & \textbf{54.83}$\pm$\textbf{0.56}\% &\textbf{48.29}$\pm$\textbf{0.51}\%& \textbf{37.49}$\pm$\textbf{0.43}\%\\
        \hline \hline
        5-way 5-shot  &miniImagenet             &  Cars           & Places         & CUB          & Plantae\\ \hline
        GNN           &80.74$\pm$0.41\%           & 42.59$\pm$0.42\%  & 72.14$\pm$0.45\%  & 63.91$\pm$0.47\%& \textbf{54.52}$\pm$\textbf{0.44}\%\\
        LRP-GNN       &\textbf{82.03}$\pm$\textbf{0.40}\%  & \textbf{46.20}$\pm$\textbf{0.46}\%  & \textbf{74.45}$\pm$\textbf{0.47}\% &\textbf{64.44}$\pm$\textbf{0.48}\% & 54.46$\pm$0.46\% \\
        \hline
    \end{tabular}
    \label{tab:GNNaccuracy}
\end{table*}

\subsection{Evaluation for Explanation-Guided Training on Cross-Domain Setting}
\label{sec:experiment_singledomain}
In this section, we evaluate the performance of RN, GNN, and CAN models trained with and without explanation-guided training on CD-FSC tasks. For more comprehensive analyses, we also implement the \textbf{Transductive inference} proposed by \cite{CAN:hou2019cross}. 
Transductive inference iteratively augments the support set using the confidently classified query images during the test phase. Specifically, we first predict the label of query images with the trained model; second, we choose the query images with higher predicted scores as the candidate images. The candidate images and their predicted label are augmented to the support set. This is an iterative process. In our experiment, we implement the transductive operation for two iterations with 35 candidates for the first iteration and 70 for the second iteration, the same strategy as \cite{CAN:hou2019cross}. GNN requires a fixed number of support images, thus we implement the transductive inference on RN and CAN models.

Table \ref{tab:accuracy_single domain} and Table \ref{tab:GNNaccuracy} summarise the accuracy of the RN, CAN, and GNN models trained with and without explanation-guided training. We can observe a consistent improvement after implementing explanation-guided training on both the seen-domain and the cross-domain test sets. The results are also competitive with the recent work on LFT \cite{FeaturewiseTranslayer:tseng2020cross} which learns a noise distribution by adding feature-wise transformation layers to the image encoder while explanation-guided training does not introduce more training parameters. To show that our approach exploits a different mechanism to improve the model, we also combine the LFT and our explanation-guided training in the next section.

\begin{table*}[tb]
    \centering
    \caption{The results of multiple domains experiment using RelationNet. We report the average accuracy of over 2000 episodes with 95\% confidence intervals. \textbf{FT} and \textbf{LFT} indicate the feature-wise transformation layer with fixed or trainable parameters. \textbf{LRP-} means explanation-guided training using LRP. \textbf{LFT-LRP} is the combination of LFT and explanation-guided training.}
    \begin{tabular}{c c c c c}
    \hline
        5-way 1-shot       &  Cars           & Places         & CUB          & Plantae\\ \hline
        RN         & 29.40$\pm$0.33\%  & 48.05$\pm$0.46\% &44.33$\pm$0.43\%& 34.57$\pm$0.38\%\\
        FT-RN      & 30.09$\pm$0.36\%  & 48.12$\pm$0.45\% &44.87$\pm$0.44\%& 35.53$\pm$0.39\%\\
        LRP-RN     & 30.00$\pm$0.32\%  & 48.74$\pm$0.45\% &45.64$\pm$0.42\%& 36.04$\pm$0.38\%\\
        LFT-RN     & 30.27$\pm$0.34\%  & 48.07$\pm$0.46\% &47.35$\pm$0.44\%& 35.54$\pm$0.38\%\\
        LFT-LRP-RN & \textbf{30.68}$\pm$\textbf{0.34}\%  & \textbf{50.19}$\pm$\textbf{0.47}\% &\textbf{47.78}$\pm$\textbf{0.43}\%& \textbf{36.58}$\pm$\textbf{0.40}\%\\ \hline \hline
        5-way 5-shot       &  Cars           & Places         & CUB          & Plantae\\ \hline
        RN         & 40.01$\pm$0.37\%  & 64.56$\pm$0.40\% &62.50$\pm$0.39\%& 47.58$\pm$0.37\%\\
        FT-RN      & 40.52$\pm$0.40\%  & 64.92$\pm$0.40\% &61.87$\pm$0.39\%& 48.54$\pm$0.38\%\\
        LRP-RN     & 41.05$\pm$0.37\%  & 66.08$\pm$0.40\% &62.71$\pm$0.39\%& 48.78$\pm$0.37\%\\
        LFT-RN     & 41.51$\pm$0.39\%  & 65.35$\pm$0.40\% &64.11$\pm$0.39\%& 49.29$\pm$0.38\%\\
        LFT-LRP-RN & \textbf{42.38}$\pm$\textbf{0.40}\%  & \textbf{66.23}$\pm$\textbf{0.40}\% &\textbf{64.62}$\pm$\textbf{0.39}\%& \textbf{50.50}$\pm$\textbf{0.39}\%\\
        \hline
    \end{tabular}
    \label{tab:multi-domain accuracy}
\end{table*}

\subsection{Synergies in Combining Explanation-guided Training with Feature-wise Transformation}
To compare and to combine our idea with the LFT method, we apply the explanation-guided training to the multiple domain experiment as \cite{FeaturewiseTranslayer:tseng2020cross}. The LFT model is trained using the pseudo-seen domain and pseudo-unseen domains. In our experiment, the miniImagenet is the pseudo-seen domain. Three of the other four datasets are the pseudo-unseen domains and the model is tested on the last domain. The pseudo-unseen domains are used to train the feature-wise transformation layers and the pseudo-seen domain is used to update the other trainable parameters of the model. If the parameters of the feature-wise transformation layers are fixed, we will get the FT method that adds the noise with a fixed distribution on certain intermediate layers. 

The performance of the standard RN, the FT and LFT methods, explanation-guided training, and its combination with LFT are shown in Table \ref{tab:multi-domain accuracy}. These models are trained with the same random seed, learning rate, optimizer, and datasets. The combination of our explanation-guided training and LFT(\textbf{LFT-LRP-RN}) achieves the best accuracy. Comparing the results of \textbf{FT-RN} and \textbf{LRP-RN}, we can see explanation-guided training is even better without introducing more trainable parameters to the model.

We remark that the improvement observed when combining explanation-guided training with LFT shows that both optimize the model from different angles. This demonstrates the independence of both approaches as well as both their strength.

\subsection{Explaining the Effect of Explanation-Guided Training}
\label{sec:feature_statistics}
In this section, we provide an intuition for the improvement of FSC models by explanation-guided training.
It is known from the information bottleneck framework that training a discriminative classifier implies learning to filter irrelevant features\cite{tishby2015deep}. This compression of task-irrelevant information is also acknowledged in recent works that shed critical light on the application of the information bottleneck to deep networks\cite{saxe2019information}. There is a difference between traditional classification and few-shot classification regarding removable information. The removable information means that some intermediate feature channels related to these removable features are not activated. The traditional classification task is to classify a fixed set of classes, therefore, removing information irrelevant to these classes will not influence the discriminative capability. For example, a classifier for cat breeds will likely learn rather the features of eyes, tails, and legs than features of sofas or grass. In FSC, the classes vary across episodes. Thus, the irrelevant information of one episode can be discriminative for the next episode. Excessive information removal can be detrimental for FSC that requires generalization across new classes. This is also the reason why we obtain low accuracy on the cross-domain test sets in Table \ref{tab:accuracy_single domain}, \ref{tab:GNNaccuracy}, and \ref{tab:multi-domain accuracy}.

From a classifier trained to classify a fixed set of classes, one would expect that in higher-layer feature maps, a few channels are highly activated somewhere in the spatial dimensions, while most channels show only low values overall. Explanation-guided training adopts explanation scores of the predicted class to re-weight intermediate features. If a classifier is overfitting and frequently predicts a wrong class label, then the explanation-guided training will identify the relevant features for the wrongly predicted class(\textbf{step2} in Section\ref{sec:EGT}), upscale them, and the subsequent loss minimization will penalize these upscaled features more (\textbf{step3\&4} in Section\ref{sec:EGT}). Thus, it avoids the intermediate features from being too specialized towards a fixed set of classes and achieves better generalization. 

To quantify the above intuition, we analyze the CNN encoded image features of the RN, GNN, and CAN models trained with or without explanation-guided training under the 5-way 5-shot setting, the same models as Section\ref{sec:experiment_singledomain}. We use the test set images of the four cross-domain datasets in this experiment. 
Each CNN encoded image feature has a shape $f_{CNN}\in\mathbb{R}^{C\times H\times W}$. We first perform a pooling over the spatial dimensions $[H,W]$, then compute a statistic over channels $C$, and finally average the statistics over the test images. We use the 95\% quantile for spatial pooling, resulting in a vector $f\in \mathbb{R}^C$. We do not use spatial average pooling due to the spatial sparsity of features as discriminative parts are usually present only in a small region of an image. For the same reason, median pooling would yield zeros mostly.
\begin{figure}
    \centering
    \includegraphics[width=0.48\textwidth]{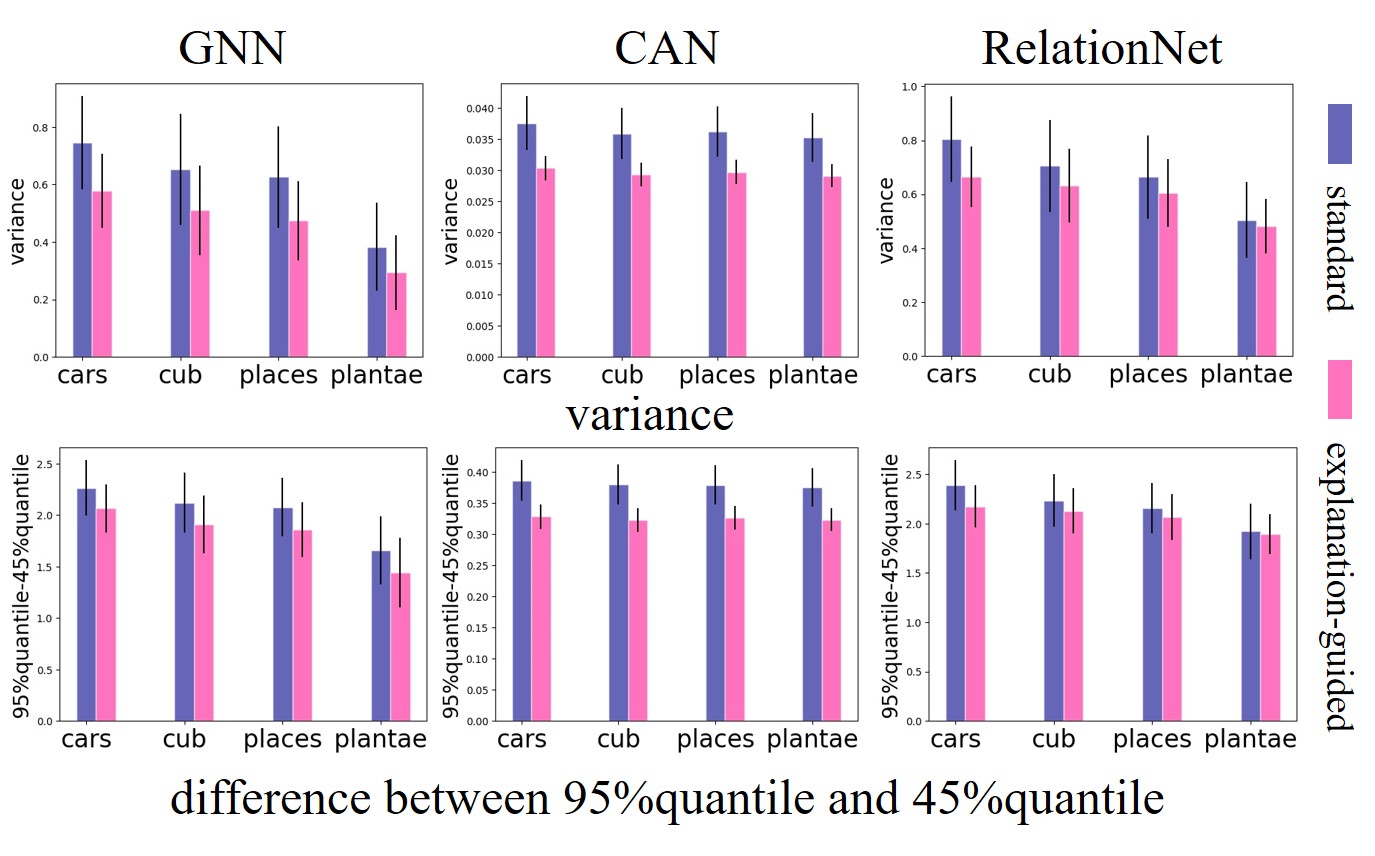}
    \caption{The variance (the first row) and quantile difference (the second row) of the CNN encoded image feature vectors. We report the mean and the standard deviation of the two feature vector statistics over all the test set images of four cross-domain datasets. The experiment models are RelationNet(RN), GNN, and CAN with(dark-pink)/without(dark-blue) explanation-guided training.}
    \label{fig:statistics_variance_quantile_diff}
\end{figure}
To verify that explanation-guided training indeed reduces excessive information removal, we observe the variance and intervals between the quantiles of the image feature vectors $f$, 
$S^2=(\sum_{i=1}^C(f^i-\Bar{f})^2)/C$ and the $95\% -45\%$ quantile difference. We calculate the two statistics for each image and calculate the mean and standard deviation of the two statistics over all the test set images of four cross-domain datasets, as illustrated in Figure \ref{fig:statistics_variance_quantile_diff}.

Lower $S^2$ and quantile difference mean that the features are not focused on a few channels but are more balanced over every channel, which preserves more diverse information and results in better generalization for new classes.
The consistent decrease of $S^2$ and quantile difference over four cross-domain datasets after applying explanation-guided training provides some evidence that the explanation-guided training effectively avoids excessive information removal and avoids overfitting on the source domain. We note that the lower $S^2$ and quantile difference are not due to lower first-order statistics such as the mean. For the CAN model, we observe an increased mean of $f$ and a decreased $S^2$ with explanation-guided training for all the cross-domain datasets.
Furthermore, the variance of some first-order statistics of $f$ over the test set also decrease with explanation-guided training. This is comparable to the effect of batch normalization, while batch normalization is naturally less effective for FSC.

\subsection{Qualitative Results of LRP Explanation for FSC Models}

The above experiments have demonstrated that, by leveraging the LRP explanation of the intermediate feature map to re-weight the same feature map, explanation-guided training effectively improves the performances of FSC models and successfully reduces the domain gap. In this section, we visualize the LRP explanation of the input images as heatmaps. From the LRP heatmaps, we can easily observe which parts of the image are used by the model to make the predictions, in other words, what features have the model learned to differentiate classes. To our best knowledge, this is the first attempt to explain the FSC models though many existing explanation methods are in principle applicable. 

\begin{figure}[tb]
    \centering
    \includegraphics[width=0.48\textwidth]{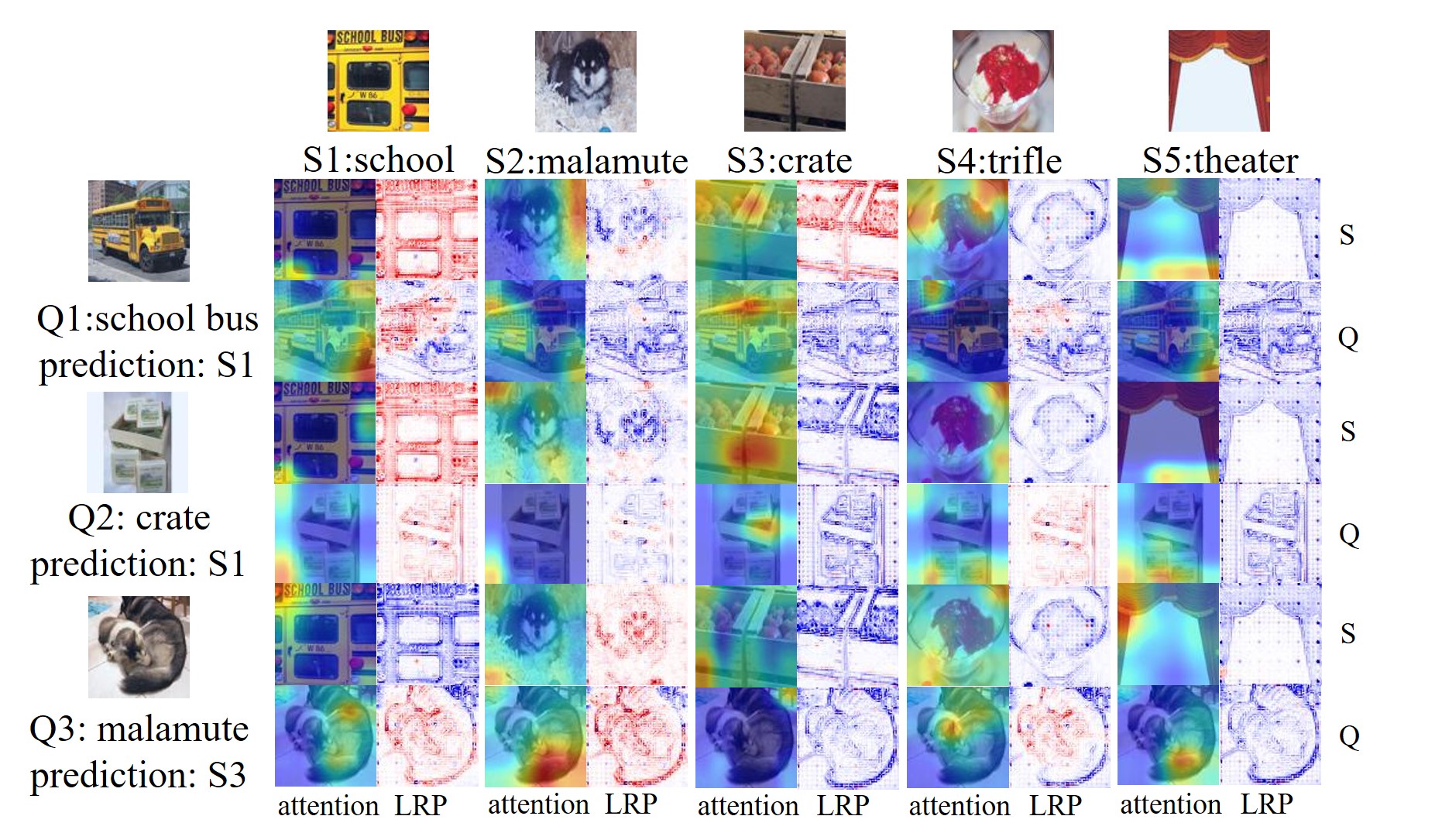}
    \caption{LRP heatmaps and the attention heatmaps of the CAN model from one episode. The model is trained under the 5-way 1-shot setting. The first row shows the support images of each class. For each query image, we illustrate the attention heatmaps and the LRP heatmaps of both the support images and the query images with 5 target labels.}
    \label{fig:visualizationCAN}
\end{figure}

Figure \ref{fig:introLRPexample} has already presented some heatmaps for the RelationNet. We further illustrate the LRP explanations of the CAN model under the 5-way 1-shot setting in Figure \ref{fig:visualizationCAN}. Since there is only one training sample per class, we also show the attention heatmaps for the support images. 
For the correctly classified $Q1$ and $Q3$, LRP heatmaps for the correct label highlight the relevant features. Specifically, the LRP heatmaps can capture the features of the window frames for the \textit{bus} and the head features for the \textit{malamute}.

On the other hand, the LRP heatmaps of the other wrong labels show more negative evidence, while we can still find some interesting resemblance between the query image and the explained label. For example, in Figure \ref{fig:introLRPexample}, when we explain the label \textit{lion} for \textit{Q1:African hunting dog}, the LRP heatmap highlights the legs of the \textit{African hunting dog} and when we explain the label \textit{cuirass} (a kind of medieval soldiers' armor) for \textit{Q2:lion}, the LRP heatmap emphasizes the round contour that resembles an armor plate. In Figure \ref{fig:visualizationCAN}, when we explain the label \textit{trifle} for \textit{Q3:malamute}, the LRP heatmap highlights the texture within a circle structure.

\section{Conclusion}
This paper shows the usefulness of explanation methods for few-shot learning during the training phase, exemplified by, but not limited to LRP. We find two points noteworthy.
Firstly, explanation-guided training successfully addresses the domain shift problem in few-shot learning, as demonstrated in the cross-domain few-shot classification task. Secondly, when combining explanation-guided training with feature-wise transformation, the model performance is further improved, indicating that these two approaches optimize the model in a non-overlapping manner. We conclude that applying explanation methods to the few-shot classification can not only provide intuitive and informative visualizations but can also be used to improve the models.

\section{Acknowledgement}
This work was supported by the Singaporean Ministry of Education Tier2 Grant MOE-T2-2-154 and the SUTD internal grant SGPAIRS1811.
This work was also partly supported by the German Ministry for Education and Research as BIFOLD (ref.\ 01IS18025A and ref.\ 01IS18037A),
and TraMeExCo (ref.\ 01IS18056A).

\bibliographystyle{IEEEtran}
\bibliography{IEEEfull}


\end{document}